\documentclass[runningheads]{llncs}
\usepackage{graphicx}
\usepackage[T1]{fontenc} 
\usepackage[misc]{ifsym}
\usepackage{hyperref}
\usepackage{url}

\usepackage{amsfonts}
\usepackage{nicefrac}

\usepackage{amsmath}
\usepackage{subfig}
\usepackage{graphicx}
\usepackage{cite}

\usepackage{microtype}
\usepackage{graphicx}
\usepackage{booktabs} 
\usepackage{multirow}
\usepackage{xcolor}

\usepackage{hyperref}

\usepackage{amsmath,amsfonts,amssymb,dsfont}

\def\R{\mathbb{R}}

\def\diag{\mathrm{diag}}

\def\our{DMFA}

\begin{document}
%
\title{Estimating conditional density of missing values using deep Gaussian mixture model\thanks{A preliminary version of this paper appeared as an extended abstract \cite{przewiezlikowskiestimating} at the ICML Workshop on The Art of Learning with Missing Values.}}
%
\titlerunning{Estimating conditional density of missing values using deep GMM}
%
\author{Marcin Przewi\k{e}\'zlikowski\inst{1} \and
Marek \'Smieja[\Letter]\inst{1}\orcidID{0000-0003-2027-4132} \and
\L{}ukasz Struski\inst{1}\orcidID{0000-0003-4006-356X}}
\authorrunning{M. Przewi\k{e}\'zlikowski et al.}
%
\institute{Faculty of Mathematics and Computer Science,\\ Jagiellonian University, Krak\'ow, Poland\\
\email{m.przewie@gmail.com}, \email{\{marek.smieja, lukasz.struski\}@uj.edu.pl}}
\maketitle              
\begin{abstract}
We consider the problem of estimating the conditional probability distribution of missing values given the observed ones. We propose an approach, which combines the flexibility of deep neural networks with the simplicity of Gaussian mixture models (GMMs). Given an incomplete data point, our neural network returns the parameters of Gaussian distribution (in the form of Factor Analyzers model) representing the corresponding conditional density. We experimentally verify that our model provides better log-likelihood than conditional GMM trained in a typical way. Moreover, imputation obtained by replacing missing values using the mean vector of our model looks visually plausible.
\keywords{missing data \and density estimation \and imputation \and Gaussian mixture model \and neural networks.}
\end{abstract}
%
%
%

\section{Introduction}

Estimating missing values from incomplete observations is one of the basic problems in machine learning and data analysis \cite{goodfellow2016deep}. A typical approach relies on replacing missing values with a single vector based on available information contained in observed inputs \cite{jerez2010missing, van2018flexible}. While imputation techniques are frequently used by practitioners, they only give point estimate instead of a probability distribution.
Quantifying the probability distribution of missing values plays an important role in generative models \cite{li2019misgan}, uncertainty prediction \cite{ghahramani1994supervised} as well as is useful in applying classification models to incomplete data \cite{smieja2018processing, williams2005analytical, dick2008learning}.

\begin{figure}[ht]
    \centering
     \includegraphics[width=0.9\textwidth]{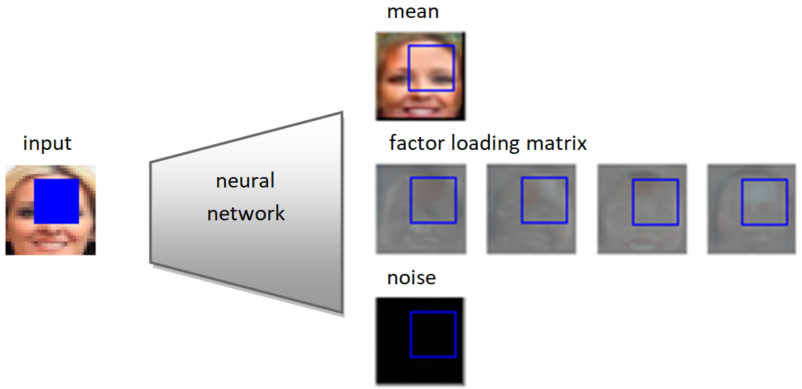}
    \caption{The idea of the proposed \our{}. Given a missing data point, our model returns the parameters of conditional Gaussian density: mean, factor loading matrix (we use 4 factors) and noise matrix for the model of Factor Analyzers, which describes a distribution of missing data (the area inside the blue square).}
    \label{fig:intro}
\end{figure}

While deep generative models such as VAE, GAN or WAE \cite{VAE, Goodfellow, WAE} are capable of modeling the distribution of complex high dimensional data, such as images, it may be difficult to use them to estimate the uncertainty contained in missing data due to the nonlinear form of decoder (generator) \cite{nazabal2020handling}. The authors of \cite{rezende2014stochastic, mattei2018leveraging} define a sampling procedure based on pseudo-Gibbs sampling and Metropolis-within-Gibbs algorithm for filling missing values by iterative auto-encoding of incomplete data. \'Smieja et al. \cite{smieja2020ae, figueiredo2020can} propose iterative algorithm for maximizing conditional density based on the dynamics of auto-encoder models. Mattei and Frellsen use importance sampling for training VAE on incomplete data as well as for replacing missing values by single or multiple imputation \cite{mattei2019miwae}. Flow models can also be trained to represent a conditional density as a neural network transformation \cite{trippe2018conditional, li2019flow}. Nevertheless, the constructed density cannot be maximized analytically. One can only produce samples or attempt to maximize the corresponding density numerically. 

In the case of shallow density models, such Gaussian mixture models (GMMs), we can easily calculate a conditional density function related to missing values in a closed-form \cite{ghahramani1994supervised, delalleau2012efficient} as well as to maximize it analytically. Moreover, simple Gaussian form of the conditional density function allows us to combine conditional GMM with other machine learning techniques that can process missing data without using any imputations at preprocessing stage \cite{smieja2018processing, smieja2019generalized}. Another related line of work has explored autoregressive models for conditional data generation or density estimation \cite{van2016conditional}.


In this paper, we propose \our{} (Deep Mixture of Factor Analyzers) for estimating the probability density function of missing values, which combines the features of deep learning models and GMMs. We construct a neural network, which takes an incomplete data point and returns the parameters of Gaussian density (represented as Factor Analyzers model) modeling the distribution of missing values, see Figure \ref{fig:intro}. Since the proposed network returns an individual Gaussian density for every missing data point, its expressiveness is higher than using GMM with a fixed number of components. In contrast to classical GMM, which estimates a density of the whole data, \our{} directly maximizes the likelihood function on missing values. In consequence, the obtained Gaussian density has a better quality in the context of missing data than the conditional distribution computed from GMM. Nonetheless, our model still provides an analytical formula for a distribution of missing values, which is useful in diverse applications, and may be more attractive than adapting deep generative models to the case of missing data. Our work is strictly related with \cite{bishop1994mixture}, but instead of using isotropic covariance matrix and many Gaussian components for conditional density, we follow \cite{richardson2018gans} and employ Factor Analyzers model, which suits better to high dimensional spaces such as images. Our preliminary work suggests that isotropic covariance is not able to model dependencies between pixels while the mixture often tends to collapse to a single Gaussian. 

Experiments conducted on image datasets confirm that the proposed \our{} gives a better value of log-likelihood function on missing values than conditional GMMs \cite{richardson2018gans}. Moreover, imputations obtained by replacing missing values with the mean vector of returned Gaussian density look visually plausible. The paper also contains a visualization of produced density function, which gives an insight into the proposed \our{}.

\section{Density model for missing data}

In this section, we introduce \our{} model. First, we recall basic facts concerning GMM and MFA in high dimensional data. Next, we show how to compute conditional density from GMM. Finally, we present the proposed \our{} -- a deep learning model for estimating conditional Gaussian density on missing values.


\subsection{Gaussian mixture model for high dimensional data}

GMM is one of the most popular probabilistic models for describing a density of data \cite{mclachlan2004finite}. A density function of GMM is given by:
$$
p(x) = \sum_{i=1}^k p_i N(\mu_i, \Sigma_i)(x),
$$
where $p_i > 0$ is the weight of $i$-th Gaussian component with mean vector $\mu_i$ and covariance matrix $\Sigma_i$ (we have $\sum_{i=1}^k p_i = 1$). Given a dataset $X \subset \R^n$, GMM is estimated by minimizing the negative log-likelihood:
$$
l(x) = -\sum_{x \in X} \log p(x).
$$

While theoretically GMM can be estimated using EM or SGD, this procedure may fail in the case of high dimensional data, such as images. Observe that for color images of size $32 \times 32$, the covariance matrix of a single component has $4.7 \cdot 10^6$ free parameters. In training phase, we need to store and invert these covariance matrices, which is computationally inefficient and may cause many numerical problems \cite{richardson2018gans}.

It is widely believed that high-dimensional data, such as images, are embedded in a lower-dimensional manifold and using full covariance matrix may not be necessary. For this reason, it is recommended to use the Mixture of Factor Analyzers (MFA) \cite{ghahramani1996algorithm} or Probabilistic PCA (PPCA) \cite{tipping1999probabilistic}, in which every Gaussian density is spanned on a lower-dimensional subspace. In contrast to the typical GMM, the covariance matrix in MFA is given by $\Sigma = AA^T + D$, where $A = A_{n \times l}$ is a factor loading matrix, which is composed of $l$ vectors $a^1,\ldots,a^l \in \R^n$ such that $l \ll n$, and $D = D_{n \times n} = \diag(d)$ is a diagonal matrix representing noise\footnote{PPCA uses spherical matrix $D$.} defined by $d \in \R^n$. The set of vectors $a^i$ defines a linear subspace, which spans a Gaussian density $N(\mu,\Sigma)$, while adding a noise matrix guarantees that $\Sigma$ is invertible. The use of MFA drastically reduces the number of parameters in a covariance matrix as well as avoids problems with inverting large matrices. Recent studies show that MFA can be effectively estimated from image data and is able to describe a higher spectrum of data density than GAN models, see \cite{richardson2018gans} for details.

\subsection{Conditional Gaussian density}

It is imporant to note that GMM can not only describe a density of data, but is also useful for quantifying the uncertainty of missing data. A missing data point is denoted by $x = (x_o,x_m)$, where $x_o$ represents known values, while $x_m$ describes absent attributes. Given a missing data point $x$, a natural question is: \emph{what is the distribution of missing values given the observed ones?} In the case of density models, the answer is given by a conditional density \cite{ghahramani1994supervised}:
$$
p(x_m | x_o) = \frac{p(x_o, x_m)}{p(x_o)} = \frac{p(x)}{p(x_o)}.
$$

In contrast to many deep generative models, e.g. GANs or VAE, the formula for conditional density can be found analytically for GMM. For a single Gaussian density $N(\mu,\Sigma)$ with $\mu=\begin{pmatrix}
  \mu_{o}\\
  \mu_{m}
\end{pmatrix}$ 
and $\Sigma = 
\begin{pmatrix}
  \Sigma_{oo} & \Sigma_{om}\\
  \Sigma_{om}^T & \Sigma_{mm}
\end{pmatrix}$, 
the conditional Gaussian density is given by:
$$
p(x_m|x_o) = N(\hat{\mu}_m, \hat{\Sigma}_m),
$$
where 
\begin{align} \label{eq:cond}
\begin{split}
&\hat{\mu}_m = \mu_m + \Sigma_{om} \Sigma_{oo}^{-1} (x_o - \mu_o),\\
&\hat{\Sigma}_m = \Sigma_{mm} - \Sigma_{om} \Sigma_{oo}^{-1} \Sigma_{om}^T.
\end{split}
\end{align}
Note that $N(\hat{\mu}_m, \hat{\Sigma}_m)$ is a Gaussian density in a lower dimensional space, where the dimension equals the number of missing values. 

To extend these formulas to the mixture of Gaussian densities, we need to find a conditional density of every Gaussian component and recalculate the weights $p_i$. Since the tails of Gaussian densities decrease exponentially, the resulting conditional GMM in high dimensional spaces typically reduces to a single Gaussian. Other components become irrelevant, because their weights (in the conditional density) are close to zero.


\subsection{Deep conditional Gaussian density for missing data}

An important advantage of GMM is that the conditional densities can be calculated and maximized analytically, which may be appealing in the context of missing data. However, GMM is not trained to estimate a density of missing data -- its objective is the log-likelihood computed on all data points. In consequence, there are no guarantees that the resulting conditional density gives optimal log-likelihood for missing values.

In this paper, we are motivated by typical deep learning models used for image inpainting \cite{iizuka2017globally, yu2018generative}. Let us recall that context-encoder \cite{pathak2016context} first generates missing values by selecting random masks for images in every mini-batch. Next, missing values are replaced by zeros and such images together with corresponding masks are processed by the auto-encoder neural network. The model is trained to minimize the mean-square error on missing values. Since the loss covers only the missing part, the context-encoder should find a better replacement for missing values than the model, which is trained to reconstruct the whole image.

Following the above motivation, the proposed \our{} creates a Gaussian density, which minimizes the negative log-likelihood on missing values. More precisely, given a data point $x \in \R^n$, we first generate a random binary mask $M$ to simulate missing attributes. The pair $(x,M)$ induces a missing data point $(x_o,x_m)$. \our{} defines a neural network $f$, which takes $(x_o,x_m)$ together with $M$ and returns the parameters of conditional Gaussian density $p(x_m|x_o)$. Following MFA model, we represent covariance matrix using factor loading matrix $A = A_{n \times l} = (a^1,\ldots,a^l)$, and the noise matrix $D = D_{n \times n}= \diag(d)$. In the case of images, $f$ simply returns the mean image $\mu$ and the covariance matrix $\Sigma = AA^T + D$ represented by $l$ images spanning a Gaussian density supplied with the noise image ($l+2$ images in total) . 

Given the output $\mu$ and $\Sigma$ of the neural network $f$, we define a conditional Gaussian density as
$$
p(x_m|x_o) = N(\mu_m,\Sigma_{mm}),
$$
where $\mu_m$ and $\Sigma_{mm}$ denote the restrictions of $\mu$ and $\Sigma$ to missing coordinates, see Figure \ref{fig:intro} for illustration.
Since the number of missing values can be different for subsequent data points, $f$ has to output the parameters of $n$-dimensional Gaussian density $N(\mu,\Sigma)$. However, $N(\mu,\Sigma)$ does not need to estimate a density of the whole data. In consequence, we do not have to use the formulas for conditional density \eqref{eq:cond}, but we can simply restrict $\mu$ and $\Sigma$ to missing attributes in order to define a conditional density $p(x_m|x_o)$. In our case, $\Sigma_{mm} = A_{m \cdot} A_{m \cdot}^T + D_{mm}$, where $A_{m\cdot}$ denotes the restriction of matrix $A = A_{n \times l}$ to the rows indexed by $m$.

\our{} is trained to minimize the negative log-likelihood of conditional density $p(x_m|x_o)$  which is given by:
$$
l(x_o,x_m) = -\log p(x_m|x_o)  = -\log N(\mu_m,\Sigma_{mm})(x_m).
$$
Observe that the above objective is calculated only on the parameters of $\mu,\Sigma$ corresponding to missing values (other entries are not used by the model). This means that $f$ can theoretically return irrelevant values on coordinates related to the observed values. The most important thing is that \our{} directly minimizes the log-likelihood of $p(x_m|x_o)$ and thus should provide a better estimate of missing values than using conditional density obtained by a typical GMM.

Let us highlight that we do not need to specify the number of mixture components as in the classical GMM. Once the neural network is fed with a missing data point, it generates an individual density for this data point. In the case of the classical mixture model, conditional density is formed by restricting the most probable Gaussian components (from the set of mixture components) to missing values. In consequence, our conditional density should be more expressive than the one obtained from the classical GMM, where the number of components is fixed.

\section{Experiments}

In this section, we compare the quality of a density produced by \our{} with a conditional density obtained from GMM. For this purpose, we use three typical image datasets: MNIST \cite{lecun1998gradient}, Fashion-MNIST \cite{xiao2017fashion} and CelebA \cite{liu2015faceattributes}.

\subsection{Gray-scale images} 

First, we consider two datasets of gray-scale images: MNIST  and Fashion-MNIST. For each test image of the size $28 \times 28$, we drop a patch of size $14 \times 14$, at a (uniformly) random location. \our{} is instantiated using 4 convolutional layers. This is followed by a dense layer, which produces the final output vectors (the number of latent dimensions $l$ determining the covariance matrix equals 4). Our model is trained with a learning rate $4\cdot 10^{-5}$ for 50 epochs. As a baseline, we use the implementation of MFA \cite{richardson2018gans} trained in a classical way\footnote{The code was taken from \url{https://github.com/eitanrich/torch-mfa}.}. The number of components $k=50$ and the number of latent dimensions $l=6$ in every Gaussian following the authors' code.

\begin{figure}[t]
    \centering
     \includegraphics[height=10cm]{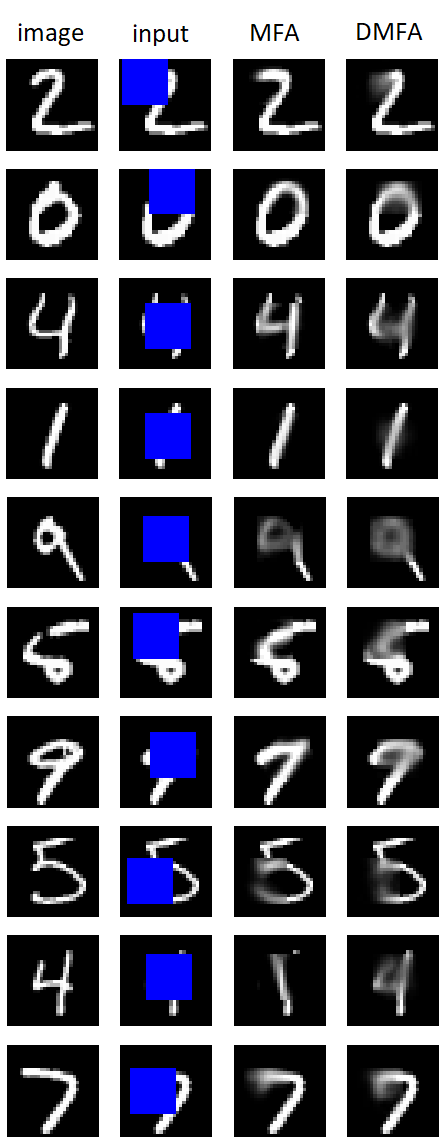} \quad
     \includegraphics[height=10cm]{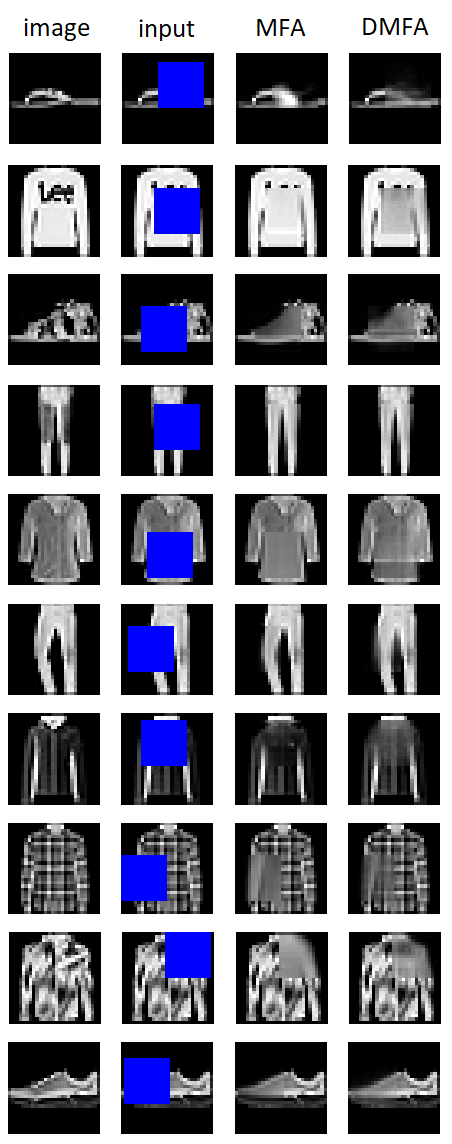}
    \caption{Sample imputation results produced by \our{} and MFA on MNIST (left) and Fashion-MNIST (right) datasets.}
    \label{fig:mnist}
\end{figure}


We examine the imputation constructed by replacing missing values with the mean vector of corresponding conditional density. Sample results presented in Figure \ref{fig:mnist} for MNIST show that MFA produces sharper imputations than \our{}. However, the results returned by MFA do not always agree with ground-truth (7th and 9th rows). This is confirmed by verifying the mean-square error (MSE) of imputations, Table \ref{tab:res}. Since \our{} usually gives images more similar to the ground-truth, it obtains lower MSE values than MFA. It is also evident from Table \ref{tab:res} that a density returned by \our{} has significantly higher log-likelihood, which means that \our{} finds a better solution to the underlying problem. 

It is worth noting that the imputation generated by MFA is in fact similar to nearest neighbor imputation. Indeed, we first select a Gaussian density which has the highest conditional probability and next project its mean vector onto the linear subspace corresponding to the missing data point (with respect to the covariance matrix). Replacing missing values using nearest neighbor usually gives sharp results, but may completely disagree with true values on the missing region. On the other hand, more blurry image corresponding to the mean vector of conditional density produced by \our{} may suggest that \our{} focuses on estimating a high-quality density function rather than finding a single value for imputation. This hypothesis is supported by high values of log-likelihood function in Table \ref{tab:res}.

Imputations generated for Fashion-MNIST show that MFA does not pay too much attention to details. While it reflects the shape of cloth items reasonably well, it is not able to predict a texture at all (compare 2nd, 5th 8th and 9th rows). On the other hand, while \our{} sometimes gives blurry results, it is more effective at discovering more detailed description of the texture. It may be caused by the fact that \our{} does not fix the number of components and returns an individual conditional density for every input image using a neural network. In consequence, its expressiveness is significantly better than MFA. While MSE values of both models are similar for Fashion-MNIST, a disproportion between log-likelihoods is again huge.

\begin{table}[t]
\setlength{\tabcolsep}{6pt}
\caption{Negative log-likelihood (NLL) and mean-square error (MSE) of the most probable imputation obtained by \our{} and MFA (lower is better).} \label{tab:res}
\vspace{0.3cm}
\centering
\begin{tabular}{ccccc}
\toprule
\multirow{2}{*}{\bf Dataset} & \multirow{2}{*}{\bf Measure} & \multirow{2}{*}{\bf MFA} & \multirow{2}{*}{\bf \our{}} & {\bf \our{}} \\
&&&& {\bf full-conv}\\
\midrule
\multirow{2}{*}{MNIST}& NLL & 58.10 & -244.81 & -\\
    & MSE & 18.59 & 12.96 & -\\
\midrule
\multirow{2}{*}{Fashion-MNIST} & NLL & -85.15 & -252.49 & -\\
    & MSE & 6.12 & 6.03 & -\\
\midrule
\multirow{2}{*}{CelebA}& NLL & -882.54 & -1222.85 & -1325.13\\
    & MSE & 9.82 & 7.73 & 4.14\\
\midrule
\end{tabular}
\end{table}

\subsection{CelebA dataset} 

We also use the CelebA dataset (aligned, cropped and resized to 32x32), which is composed of color face images, with missing regions of size $16 \times 16$. Processing of CelebA images is more resource demanding than working with MNIST and Fashion-MNIST datasets. Therefore, in addition to the convolutional neural network with a dense layer from the previous example, we also examine a fully-convolutional neural network based on DCGAN \cite{radford2015unsupervised}, which does not contain any dense layer and, in consequence, suits better to large data. Our preliminary experiments suggested that it is difficult for the fully-convolutional model to find a good candidate for the mean vector of returned density from scratch. To cope with this problem, we supply the negative log-likelihood with MSE loss\footnote{In fact, minimizing MSE leads to fitting a Gaussian density with isotropic covariance, so this form of loss function still optimizes a log-likelihood.} for the first 10 epochs, which is later turned off. Again, we put $l=4$ and train \our{} with a learning rate $4\cdot 10^{-5}$ for 50 epochs. The baseline MFA model uses $k=300$ components and $l=10$ latent dimensions.

\begin{figure*}[t]
    \centering
     \includegraphics[height=6cm]{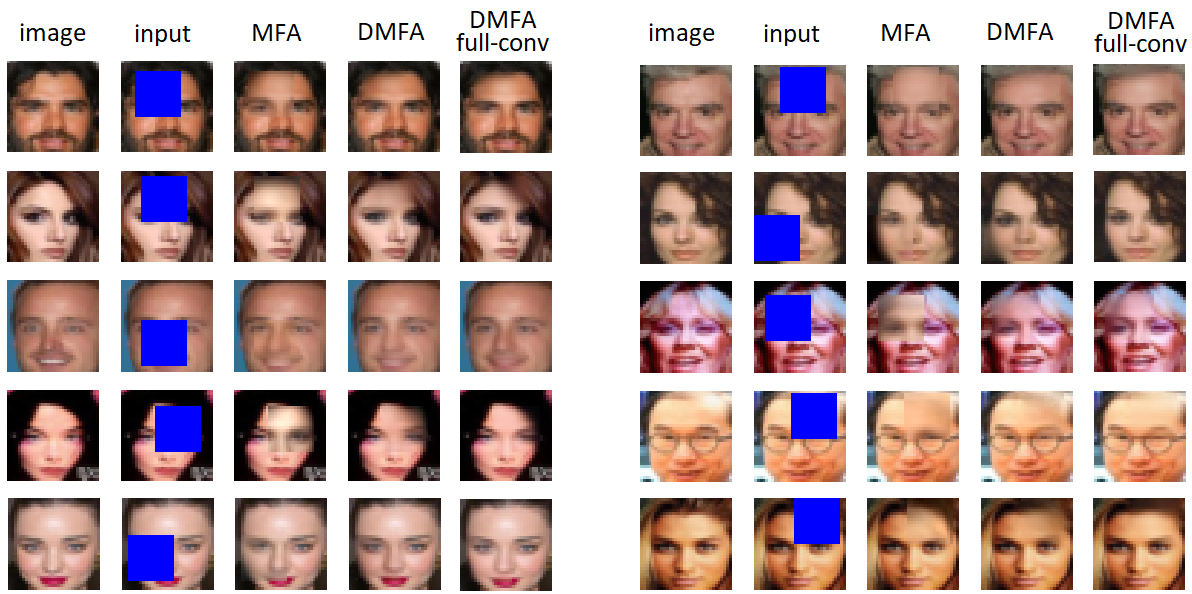}
    \caption{Sample imputation results produced by MFA and two versions of \our{} model on CelebA.}
    \label{fig:celeba}
\end{figure*}

The Figure \ref{fig:celeba} shows that the fully convolutional version of \our{} leads to the best looking imputations (last column). The second version of \our{} also coincides with ground-truth, but its quality is worse. The results obtained by MFA are not satisfactory. Quantitative assessment, Table \ref{tab:res}, confirms that \our{} implemented using fully convolutional network outperforms standard \our{} both in terms of MSE and log-likelihood.

\subsection{Parameters of conditional density}

\begin{figure}[h]
    \centering
     \includegraphics[height=12cm]{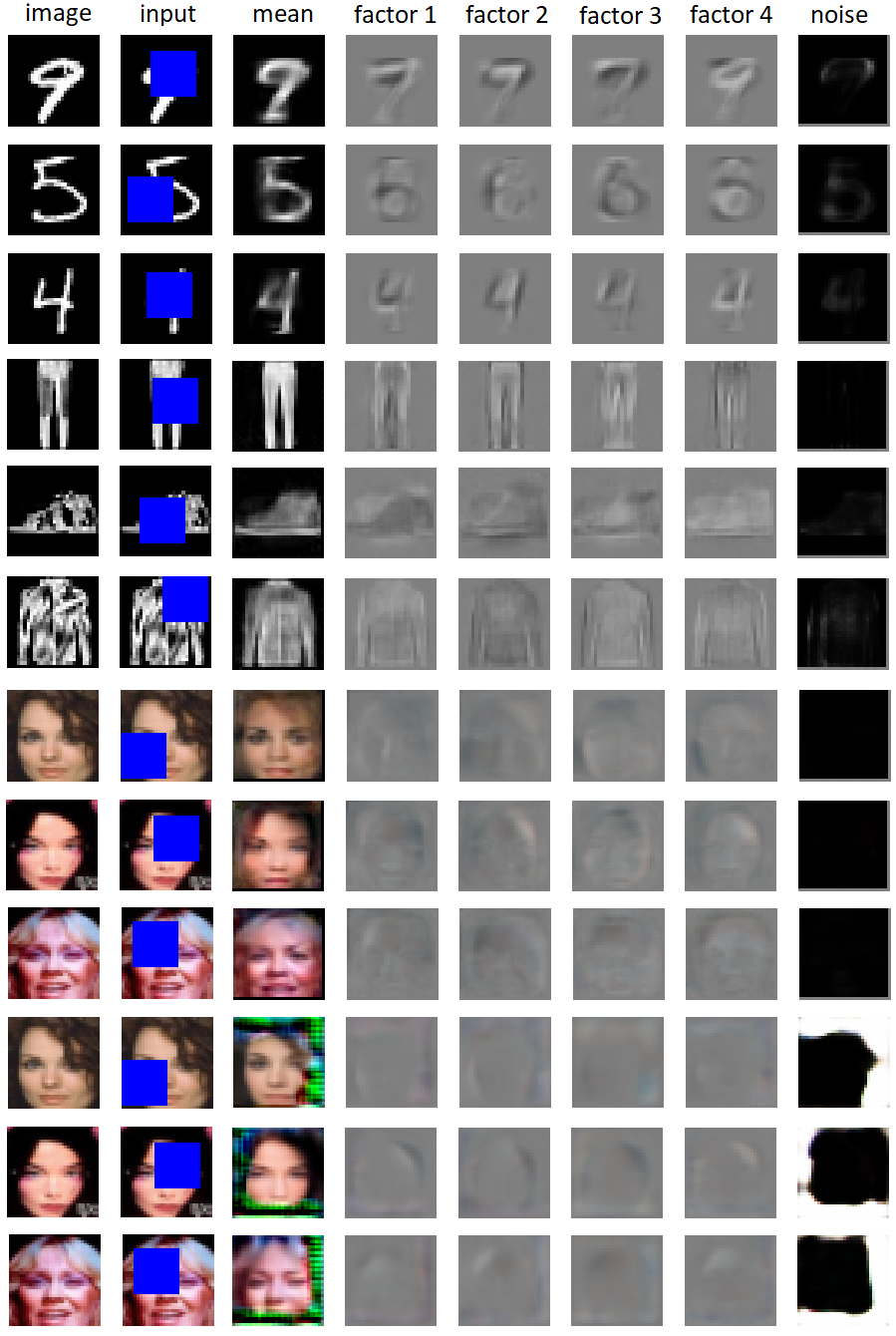} \quad
    \caption{Parameters of Gaussian distribution returned by \our{} (the last three rows correspond to the fully convolutional architecture).}
    \label{fig:params}
\end{figure}

Finally, we analyze a density estimated by \our{}. Figure \ref{fig:params} shows images corresponding to the mean vector, the factors determining the covariance matrix and the noise vector. 

Note that \our{} returns the parameters of $n$-dimensional Gaussian density, but the conditional density is obtained by restricting them to missing attributes. Interestingly that the model with an additional dense layer (1st-9th rows) gives a reasonable estimate on the whole image. Note however that the mean vector outside the mask is not exactly the same as the input data -- it is especially evident for CelebA datasets. Introducing a dense layer allows the neural network to easily fuse the information from the whole image, which may help the neural network to fit better to the changing area of missing data. On the other hand, it is evident that fully convolutional architecture focuses only on predicting values at missing coordinates (and small area that surrounds it). It is generally difficult (or even impossible) to fully convolutional networks to mix the information from distant areas of the image and thus it concentrates only on estimating a density on the required missing region.

It is evident that the factors determining the covariance matrix contain diverse shapes, which allows the obtained density to cover a wide spectrum of possible values. For example, the first three factors in the first row correspond to digit "7" while the last one is more similar to the digit "9". Factors in the third row determine different writing styles of digit "4". In the case of Fashion-MNIST, factors are mainly responsible for adding brightness intensity to a given shape. Observe that the factors for fully convolutional architecture have lower variance than using additional dense layer. It is partially confirmed by lower MSE and negative log-likelihood values. The magnitude of the noise (last column) is very low (except for MNIST), which is a positive effect, because the noise is added only to guarantee the invertibility of covariance matrix.

\section{Conclusion and future work}

We proposed a deep learning approach for estimating the conditional Gaussian density of missing values given the observed ones. Experiments showed that the obtained density has significantly lower value of negative log-likelihood function than conditional GMM trained in a classical way. Moreover, imputations produced by replacing missing values with the mean vector of resulting Gaussian look visually plausible.

In the future, we will use \our{} in a combination with machine learning approaches dealing with missing data. In particular, we plan to apply the obtained conditional density to general classification neural networks, which do not need to fill in missing values at the preprocessing stage, but can process incomplete data using a Gaussian estimate of missing values. We would also like to examine \our{} on higher resolution images. Moreover, we will focus on designing a strategy for training \our{} on incomplete data.

\section*{Acknowledgements}

The work of M. \'Smieja was supported by the National Science Centre (Poland) grant no. 2018/31/B/ST6/00993. The work of \L{}. Struski was supported by the National Science Centre (Poland) grant no. 2017/25/B/ST6/01271 as well as the Foundation for Polish Science Grant No. POIR.04.04.00-00-14DE/18-00 co-financed by the European Union under the European Regional Development Fund.

\bibliographystyle{splncs04.bst}
\bibliography{paper}
\end{document}